\documentclass[conference]{IEEEtran}
\IEEEoverridecommandlockouts
\usepackage{amsmath,amssymb,amsfonts}
\usepackage{algorithmic}
\usepackage{graphicx}
\usepackage{textcomp}
\usepackage{xcolor}
\usepackage{booktabs}
\usepackage{cite}
\usepackage{url}

\begin{document}

\title{Compressed Recurrent Feedback in Tsetlin Machines: A Reproducible Boolean-FSM Study}

\author{
\IEEEauthorblockN{Ankit Kumar\textsuperscript{1}, Utkarsh Raj\textsuperscript{1}, Rishad Shafik\textsuperscript{2}, and Sudip Roy\textsuperscript{1}}
\IEEEauthorblockA{\textsuperscript{1}Department of Computer Science and Engineering, Indian Institute of Technology Roorkee, Roorkee, India}
\IEEEauthorblockA{\textsuperscript{2}Microsystems Research Group, Newcastle University, Newcastle upon Tyne, UK}
}

\maketitle

\begin{abstract}
Sequential inference on small devices requires a model to retain useful history without repeatedly processing a long input record. A Recurrent Tsetlin Machine (RTM) provides this memory by returning Boolean clause outputs from one time step as inputs to the next. Direct feedback, however, grows with the clause bank and can make the recurrent input unnecessarily wide. This paper investigates a fixed-width alternative. We combine clause activations by exclusive-OR (XOR) folding, retain the folded bits at two time scales, and threshold them back to a binary state. The resulting design reduces 480 clause activations to 96 recurrent bits. We evaluate the method on a reproducible Boolean finite-state-machine benchmark with explicit transition rules, data splits, and random seeds. Across 144 runs, the compressed model obtains $61.47 \pm 6.74\%$ and $62.94 \pm 9.92\%$ accuracy on the two task families. Raw clause feedback changes these means by less than one percentage point, while increasing the recurrent width tenfold and measured host execution time by $4.38\times$ and $3.71\times$. Gated neural models remain more accurate, and a no-feedback control retaining only short input history achieves comparable or slightly higher accuracy. On this benchmark, folding matches raw feedback within small empirical margins at a much narrower interface; these findings also underscore the critical necessity of no-feedback recurrence controls when benchmarking sequence models.
\end{abstract}


\section{Introduction}
The Tsetlin Machine (TM) is a logic-based learning algorithm built from simple finite-state automata [2]. Instead of learning dense real-valued transformations, it forms propositional clauses: conjunctions of input conditions that vote for or against each class. The resulting Boolean operations are transparent and well suited to digital implementation, which has made the TM attractive for resource-constrained edge systems.

Conventional TMs process samples independently. This is a limitation for sensor streams, event traces, and other tasks in which the meaning of the current observation depends on what happened earlier. Neural sequence models address the same problem by carrying a hidden vector through time. Recurrent neural networks (RNNs), gated recurrent units (GRUs), and long short-term memory networks (LSTMs) learn such real-valued states, but their dense arithmetic can be costly on small devices [1]. The Recurrent Tsetlin Machine (RTM) instead returns clause activity from the current step as Boolean context for the next step [3].

Clause feedback gives the RTM a direct form of temporal memory, but it also introduces a scaling problem. A larger clause bank produces a wider recurrent state, and every additional feedback bit becomes part of the next TM input. We address this problem by replacing the raw clause vector with a fixed-width summary. Clause outputs are first combined by exclusive-OR (XOR) folding. Fast and slow moving averages then retain short- and longer-term activity before thresholding restores a binary state. The complete data path is given in Section~III and Fig.~1.

Evaluating this idea requires more than a compressed-versus-uncompressed accuracy comparison. An apparent gain may come from the observed input window or from ensemble voting rather than clause recurrence. Likewise, a compact feedback pattern may support prediction without representing the task's true hidden state. For this reason, we use Boolean finite-state machines (FSMs), whose hidden transition rules are known to the experimenter but not supplied to the learner. Section~IV specifies the generator, state-to-output map, input distribution, sequence lengths, splits, and seeds. Multiple circuit and training seeds are used because state count alone does not determine difficulty and a single run gives weak evidence for a stochastic learner [4].

The main contributions are:
\begin{itemize}
\item a fixed-width recurrent interface folding 480 clause activations into 48 bits via XOR hashing at two time scales;
\item a reproducible Boolean-FSM benchmark with exported circuits, transition tables, split arrays, and file hashes; and
\item a matched evaluation covering neural baselines, component ablations, raw feedback, no-feedback controls, and probes.
\end{itemize}

These observations motivate three research questions:
\begin{itemize}
\item \textbf{RQ1 (Baselines):} How does compressed RTM compare with majority predictors and neural sequence learners?
\item \textbf{RQ2 (Components):} What are the contributions of clause feedback, ensembles, fold width, multi-view inputs, and timescales?
\item \textbf{RQ3 (Compression vs. Representation):} How do compressed and raw feedback compare in accuracy, width, runtime, and hidden-state recovery?
\end{itemize}

Across the six circuits, folding changes raw-feedback accuracy by less than one percentage point, yet the no-feedback control is slightly more accurate. We treat this as evidence that the compressed interface effectively reproduces raw-feedback performance on these tasks without requiring wide recurrent buses, while highlighting that short input histories already provide substantial temporal context.

The remainder of this paper is organized as follows: Section~II reviews relevant TM and sequence-learning literature. Sections~III and IV define the compressed RTM model and Boolean-FSM benchmark. Section~V outlines the experimental setup. Sections~VI, VII, and VIII present the empirical results, discussion, and reproducibility artifact, while Section~IX details threats to validity and Section~X concludes the paper.

\section{Related Work}
This section reviews foundational Tsetlin Machine principles, compressed edge realizations, and established neural sequence learning architectures.

\subsection{Tsetlin Learning and Model Variants}
The original TM introduced learning with propositional clauses that remain readable as Boolean rules [2]. Later work expanded the same principle to images and regression [5], [6]. Weighted clauses reduce rule duplication [7], multigranular clauses reduce sensitivity to a single specificity setting [8], and Drop Clause improves diversity and robustness [9]. Parallel architectures and convergence analyses further clarify how clause learning can scale and why its simple operators work [10], [11]. None of these methods addresses what happens when clause activity itself becomes a time-varying input to an RTM.

\subsection{Edge Realization and Temporal Context}
The bit-oriented representation has also motivated low-complexity hardware [12], compressed edge-model generation in REDRESS [13], automated system-on-chip construction in MATADOR [14], and processor support for inference [15]. These systems mainly reduce or accelerate a model whose parameters are fixed after training. The RTM creates a different cost because its clause outputs change at every step and are returned as temporal context [3]. Our work connects these two directions by reducing that dynamic interface; it does not propose a new accelerator or claim to shrink the stored TM itself.

Established neural alternatives include simple recurrent networks, gated recurrent designs, and attention-based Transformers [16]--[19]. They provide useful accuracy references when trained on the same sequences, although differences in capacity and optimization prevent a universal ranking. Representation probes require similar care [20]: decoding hidden FSM state is meaningful evidence, whereas merely observing many distinct feedback patterns is not.

Static and recurrent compression act on different objects. Static TM compression reduces the learned model through weighting, pruning, or compact automata encoding. Recurrent-interface compression reduces the data produced at one time step and returned at the next. The clause bank can remain unchanged even when the state carried between decisions becomes much smaller. This paper studies the latter and makes no claim about parameter compression.

That choice determines the controls. A feed-forward comparison would mix the effect of recurrence with the effect of the observed input window. Comparing only compressed and raw feedback would reveal the cost of folding, but not whether clause feedback helps at all. The evaluation includes both raw feedback and a no-feedback control that retains the same two-symbol history and previous prediction.

\section{Compressed RTM}
Figure~1 gives an overview of the proposed recurrent path. At time step $t$, the RTM receives observed features $d_t$ and binary context $h_{t-1}$ from the previous step. It evaluates its clauses, produces prediction $\hat{y}_t$, and exposes clause activations $c_t$:
\begin{equation}
z_t = [d_t, h_{t-1}], \quad (\hat{y}_t, c_t) = \mathrm{TM}(z_t),
\end{equation}
The model uses 240 clauses for each of two classes, giving $c_t \in \{0, 1\}^{480}$. Compression is applied only to the activity returned at the next step; the learned clause bank is unchanged. The reported cost measures are recurrent-interface width, resulting input dimension, and host time rather than learned-model storage or target-device energy.

\subsection{Binary Input and Feedback}
Each input symbol $x_t \in \{0, \dots, 15\}$ is expanded into several Boolean views: its four binary digits, 16 cumulative threshold (thermometer) bits, 16 one-hot identity bits, and six pairwise XOR bits. This produces $e(x_t) \in \{0, 1\}^{42}$. Because Tsetlin Machines operate purely on propositional logic and do not perform continuous arithmetic or internal thresholding, this multi-view expansion explicitly linearizes distinct mathematical properties into Boolean literals: thermometer bits capture monotonic ordinal thresholds ($x \ge \theta$), one-hot bits allow exact identity matching ($x = k$), binary digits provide compact modular representations, and pairwise XOR bits enable single-clause parity and difference detection. The observed input also includes the previous symbol and a one-hot encoding $p_{t-1}$ of the previous prediction, providing autoregressive label continuity. With zero initialization at sequence start,
\begin{equation}
d_t = [e(x_{t-1}), e(x_t), p_{t-1}] \in \{0, 1\}^{86}.
\end{equation}

The wide clause vector is then reduced. A seeded, fixed assignment $\pi(j)$ maps clause $j$ to one of $B = 48$ groups, called bins. Within each bin, exclusive OR returns one when an odd number of assigned clauses are active:
\begin{equation}
q_{t,k} = \bigoplus_{j:\pi(j)=k} c_{t,j}, \quad k = 1, \dots, B.
\end{equation}
XOR was selected over alternative pooling methods (such as logical OR, AND, or majority voting) because it preserves pure Boolean computation without introducing real-valued multipliers or threshold comparisons. Furthermore, OR-pooling quickly saturates to 1 when clause banks are active and AND-pooling vanishes to 0, whereas XOR operates as a balanced 1-bit parity hashing function across clause subsets. In hardware, bitwise XOR maps directly to single-cycle digital XOR trees with minimal area and propagation delay.

Two exponential moving averages (EMAs) then provide different memory speeds: the smaller coefficient responds quickly, while the larger coefficient changes more slowly:
\begin{equation}
m^{(r)}_t = \alpha_r m^{(r)}_{t-1} + (1 - \alpha_r) q_t, \quad \alpha_r \in \{0.3, 0.9\},
\end{equation}
The moving averages are converted back to Boolean values by a threshold:
\begin{equation}
b^{(r)}_t = \mathbf{1}[m^{(r)}_t \ge 0.5], \quad h_t = [b^{(0.3)}_t, b^{(0.9)}_t] \in \{0, 1\}^{96}.
\end{equation}
Together, the two memories form a 96-bit state, so the complete compressed input has $86 + 96 = 182$ bits. In the raw comparison, folding is removed and both memories operate on all 480 clause bits, producing 960 recurrent bits and a 1046-bit input.

Five members use different training orders and fold assignments, and their predictions are combined by majority vote. As shown in Fig.~1, each labeled step forms $d_t$, evaluates the TM, folds $c_t$, updates and thresholds both memories, and then applies the TM training update. This order is unchanged during training and testing.

\begin{figure}[t]
\centering
\includegraphics[width=\columnwidth]{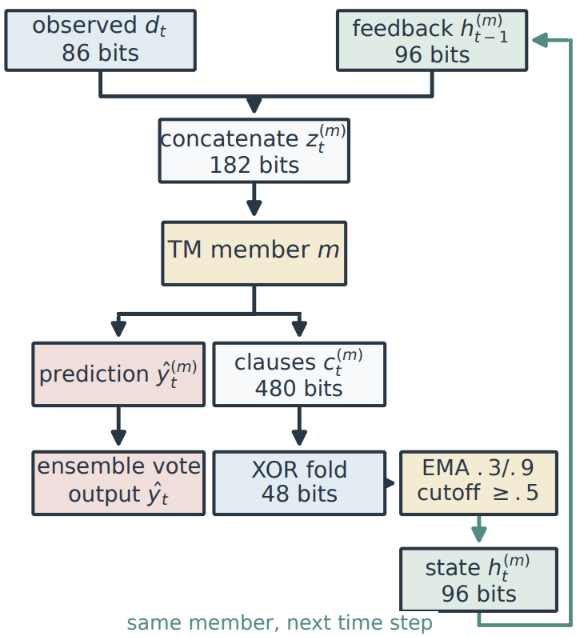}
\caption{Compressed RTM data flow. Clause activations are folded, retained through fast and slow exponential moving averages, and thresholded before returning at the next step. Five independently seeded members vote on the output.}
\label{fig:dataflow}
\end{figure}

\subsection{Causal Update Order and Interface Accounting}
The arrows in Fig.~1 also enforce causality. At sequence start, the input history, previous-prediction vector, EMA values, and recurrent bits are reset; state is never carried between independent sequences. At step $t$, a member first predicts from $[d_t, h_{t-1}]$. Only afterward are its current clause activations folded to create $h_t$ for step $t+1$. Training follows the same order, so the current prediction cannot use the target or any state derived after that target is observed.

For $B$ fold bins and $R$ retained timescales, the per-member recurrent and total input widths are
\begin{equation}
W_h = BR, \quad W_z = 86 + BR.
\end{equation}
The compressed setting uses $(B, R) = (48, 2)$, giving $(W_h, W_z) = (96, 182)$. Raw feedback substitutes the $C = 480$ clause outputs for the fold bins and retains both timescales, giving $(960, 1046)$. These are interface dimensions, not trainable floating-point parameter counts. Ensemble voting replicates the member computation but does not concatenate member states; reported widths are therefore per member, while measured host time includes all five members.

\subsection{Fold Semantics and Collision Behavior}
The assignment $\pi$ is generated from the member and projection seeds when that member is constructed, then held fixed during training and evaluation. Each of the 480 clause positions maps to exactly one bin; no learned matrix or target-dependent operation is used in the compressor. A bin is one precisely when an odd number of its active clauses map to it. Consequently, two simultaneously active clauses in the same bin cancel, and the original clause identities cannot in general be reconstructed. The two EMA memories operate on the same 48 folded bits and add temporal persistence, not additional evidence channels.

The construction bounds the recurrent interface independently of clause-bank size, but it does not compress the learned automata or reduce the TM's trainable state. Across all five ensemble members, the total static model parameter memory contains $5 \times 480 \times 2 \times W_z$ two-state automata bits. Recurrent-interface compression specifically reduces the dynamic communication bandwidth and input-register size between time steps. Different ensemble members use different fixed assignments so that voting can average over training-order and sketch variation. The evaluation isolates these choices through 32-bit, single-member, and raw-feedback variants.

\section{Reproducible Boolean-FSM Benchmark}
An FSM is a controlled process with a hidden state, an observed input, and a fixed rule that maps both to the next hidden state. This makes the true memory known to the experimenter while keeping it hidden from the learner, which is useful for testing both prediction and state recovery. Our two families use $d \in \{6, 8\}$ hidden bits, allowing at most 64 or 256 states. These labels describe state width, not task difficulty.

At step $t$, the four binary digits of input symbol $x_t$ are appended to current state $s_t$ to form source vector $u_t = [s_t, \mathrm{bin}_4(x_t)]$. A seeded generator chooses signal sources and a three-gate Boolean circuit for each next-state bit:
\begin{align}
v_{j,1} &= g_{j,1}(u_{t,a_{j,1}}, u_{t,b_{j,1}}), \nonumber \\
v_{j,k} &= g_{j,k}(v_{j,k-1}, u_{t,b_{j,k}}), \quad k \in \{2, 3\}.
\end{align}
All state bits update simultaneously as $s_{t+1,j} = v_{j,3}$. The prediction target is the parity of this updated state:
\begin{equation}
y_t = \left( \sum_{j=1}^d s_{t+1,j} \right) \bmod 2.
\end{equation}
In plain terms, the label is one when an odd number of updated state bits are one, and zero otherwise. It is not the parity of the input symbols or of an entire sequence. Figure~2 summarizes how each input and hidden state produce the next state and label.

\begin{figure}[t]
\centering
\includegraphics[width=\columnwidth]{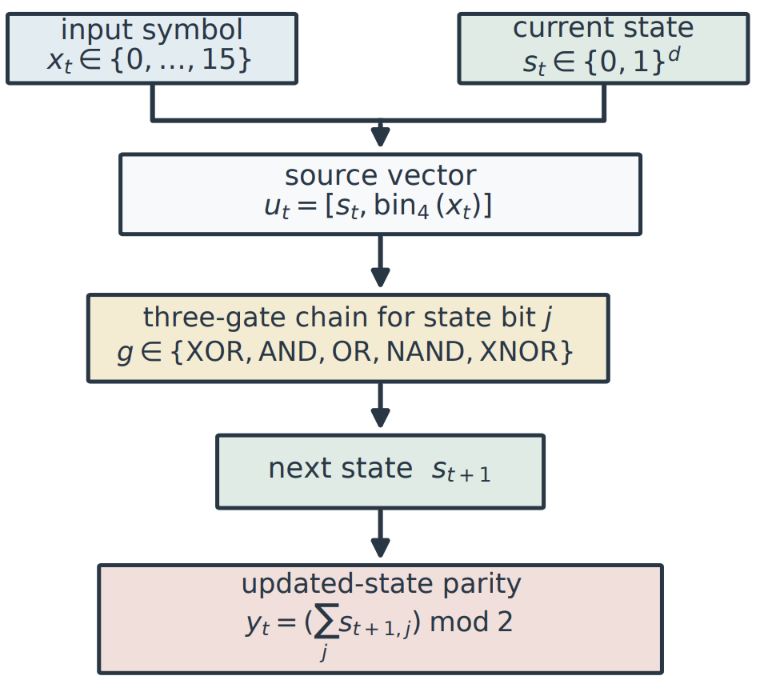}
\caption{Boolean-FSM generator. Circuit seeds fix the hidden transition rule; dataset seeds sample initial states and observed input sequences. The label is the parity of the updated hidden state.}
\label{fig:generator}
\end{figure}

Inputs and initial states are sampled independently and uniformly. Sequence lengths are 25 and 30 for the 6- and 8-bit families, and each circuit has 1000 training, 300 validation, and 500 test sequences. A fixed split seed maps to separate dataset seeds for those partitions. The exported artifact includes the arrays, circuit definitions, complete transition tables, output maps, and file hashes. The three circuit seeds per family produce the structural variation in Table~I; their different majority baselines show why a universal 50\% reference would be misleading.

The structural columns are computed exhaustively over every current-state/input pair, rather than estimated from sampled sequences. ``Distinct next'' counts unique successor states over the full transition table. Mean out-degree first counts, for each current state, how many distinct successors are produced by the 16 input symbols and then averages across states. Mean bit flips is the average Hamming distance between $s_t$ and $s_{t+1}$ over the same table. These quantities do not define a total ordering of difficulty, but they expose important differences hidden by the labels ``64 states'' and ``256 states.''

The majority baseline is reported separately for each circuit because output balance changes with both the transition function and sampled sequences. It is the accuracy of always predicting that circuit's most common test label. All learned models receive the same inputs and labels for a matched circuit. Initial hidden states are never provided to the models; they can influence prediction only through the observed sequence. The same hidden states are retained separately for the representation probe.

\section{Experimental Setup}
This section details the hyperparameter configurations, baseline neural models, ablation variants, and statistical uncertainty protocols used in our evaluation.

\subsection{Model Configurations and Baselines}
We evaluate the architecture in Section~III on the matched Boolean-FSM circuits from Section~IV. The RTM uses 240 clauses per class, voting threshold $T = 120$, clause-specificity parameter $s = 6.0$, weighted clauses, and eight training passes through the data. These hyperparameters are chosen to provide a stable, symmetric operating point: 240 clauses per class (480 total) provide adequate capacity for learning Boolean FSM logic while establishing a realistic baseline for a tenfold interface compression; $T = 120$ sets the voting threshold to half the clause bank, balancing Type~I and Type~II feedback to prevent premature saturation; and $s = 6.0$ establishes balanced inclusion probabilities across the 86-bit input literals. Accuracy is the fraction of correct labels across all test time steps. Three circuit seeds and three training seeds produce nine runs per family and variant, and 144 RTM runs in total. We report the mean and sample standard deviation; paired ablation differences use unadjusted 95\% Student-$t$ intervals. Python 3.9, NumPy 1.26.4, TMU 0.7.1, and PyTorch 2.8.0 ran on one ARM64 macOS host. Wall time includes training and test evaluation, but is not a hardware benchmark.

Table~II summarizes the model, ablation, and probe runs. Each RTM epoch shuffles complete training sequences while preserving time order within a sequence. Input history, previous prediction, and recurrent memory are reset before every sequence. Training then performs one causal prediction and one TM update per labeled step, using the update order in Fig.~1. Every RTM variant follows the same eight-epoch schedule; validation data are exported but do not select an RTM checkpoint.

Five causal baselines use the same arrays and seeds: an eight-symbol-window multilayer perceptron (MLP), a simple RNN, a GRU, an LSTM, and a Transformer [16]--[19]. All map symbols to 16-dimensional learned vectors. Recurrent models use one 32-unit layer, the MLP uses two 32-unit layers, and the Transformer uses two attention layers. AdamW uses learning rate $10^{-3}$, weight decay $10^{-4}$, batch size 32, and at most 60 epochs with eight-epoch patience. Validation selects neural checkpoints, whereas the fixed RTM schedule does not. These 90 runs compare the stated implementations on common data, not equal parameter counts or optimization budgets.

The eight RTM variants include primary comparisons (full compressed feedback, uncompressed 480-bit clause feedback, and no clause feedback) and exploratory component ablations (one ensemble member, a 32-bit fold, no thermometer features, one EMA with $\alpha = 0.5$, and instantaneous binary feedback). Each variant changes one part of Fig.~1. The no-feedback control still receives two observed symbols and the previous prediction, so it removes clause-derived state rather than all temporal information.

A nearest-centroid probe tests whether similar feedback signatures correspond to the same hidden FSM state. For each run, the probe forms a representative binary signature for every observed state from 70\% of the first member's test time points, then ranks states for the remaining 30\% by Hamming distance. If $K$ states occur, random top-1 and top-5 references are $1/K$ and $\min(5/K, 1)$. The split does not test held-out-sequence generalization, but performance below the matched random reference is sufficient to reject a direct state-identity interpretation.

\subsection{Paired Comparisons and Uncertainty}
Every variant is evaluated on the same nine circuit/training-seed combinations within a family (three circuit seeds $\times$ three training seeds nested within each circuit). Its ablation effect is the accuracy change from the full model on each matched pair. Pairing prevents the circuit differences in Table~I from being mistaken for component effects. Reported intervals summarize the mean of the nine paired changes. They are not adjusted across ablations and, with three circuit seeds, quantify variation within this design rather than characterize the full generator distribution.

\begin{table*}[t]
\caption{Circuit structure, test baseline, and full-RTM accuracy (three training seeds).}
\label{tab:circuits}
\centering
\begin{tabular}{lccccccr}
\toprule
Family & Circuit seed & Nominal states & Distinct next & Mean out-degree & Mean bit flips & Majority baseline & Full RTM \\
\midrule
6-bit & 42 & 64 & 48 & 9.12 & 2.75 & 50.45\% & 61.38\% \\
6-bit & 43 & 64 & 48 & 4.50 & 3.25 & 61.74\% & 69.29\% \\
6-bit & 44 & 64 & 64 & 6.00 & 3.00 & 50.30\% & 53.74\% \\
8-bit & 42 & 256 & 108 & 6.31 & 4.12 & 55.17\% & 76.12\% \\
8-bit & 43 & 256 & 164 & 11.81 & 4.12 & 53.56\% & 55.42\% \\
8-bit & 44 & 256 & 192 & 9.69 & 4.00 & 51.51\% & 57.29\% \\
\bottomrule
\end{tabular}
\end{table*}

\begin{table}[t]
\caption{Evaluation matrix.}
\label{tab:eval_matrix}
\centering
\resizebox{\columnwidth}{!}{%
\begin{tabular}{llr}
\toprule
Study & Factorization & Runs \\
\midrule
RTM variants & 2 families $\times$ 3 circuits $\times$ 3 seeds $\times$ 8 variants & 144 \\
Neural baselines & 2 families $\times$ 3 circuits $\times$ 3 seeds $\times$ 5 models & 90 \\
State probes & first member of all RTM variant runs & 144 \\
\bottomrule
\end{tabular}%
}
\end{table}

\begin{figure}[t]
\centering
\includegraphics[width=\columnwidth]{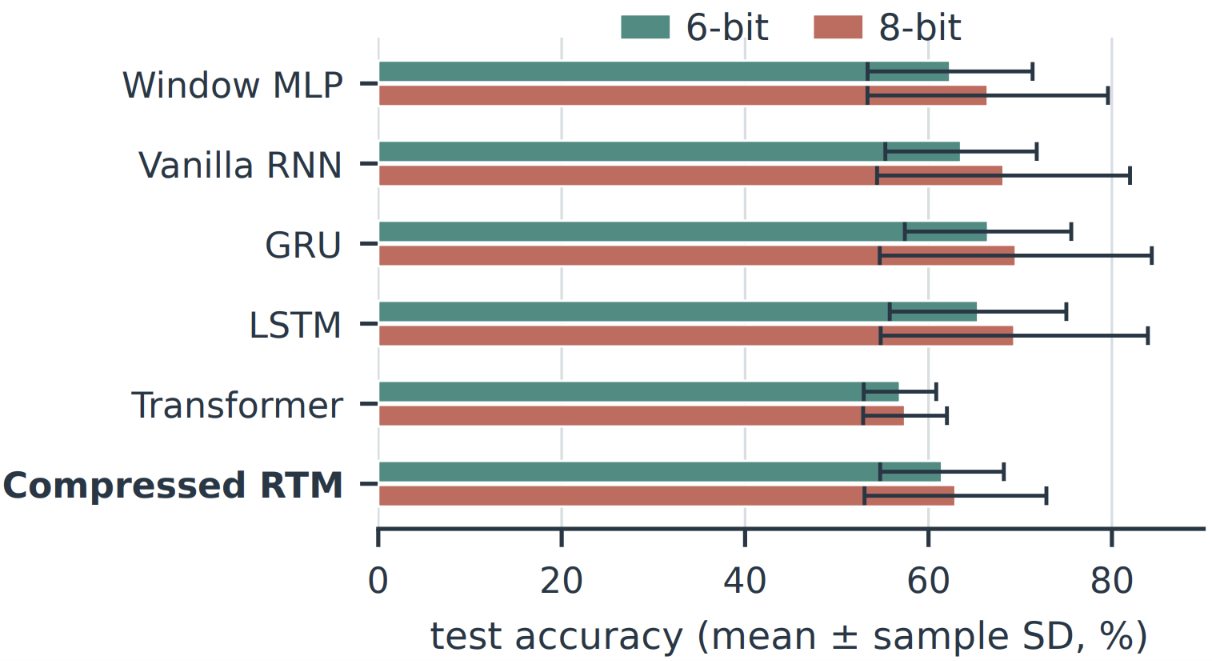}
\caption{Mean test accuracy over nine matched circuit/training-seed runs. Error bars show sample SD across those runs, combining circuit and training variation; they are not confidence intervals.}
\label{fig:neural_baselines}
\end{figure}

\begin{figure}[t]
\centering
\includegraphics[width=\columnwidth]{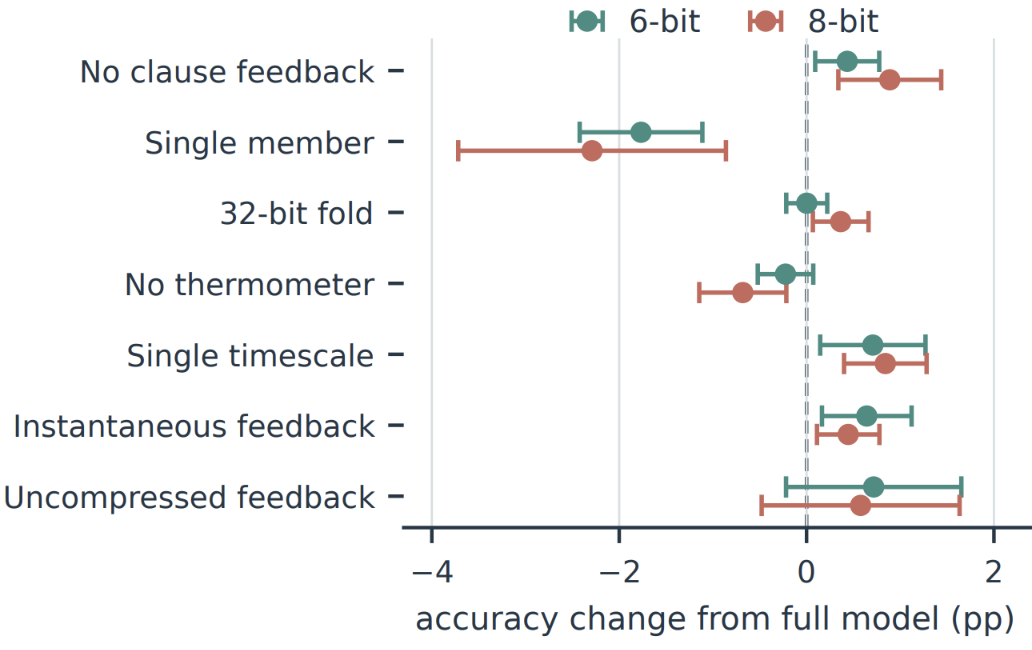}
\caption{Paired accuracy change from the full model over nine runs; bars show 95\% $t$ intervals.}
\label{fig:ablations}
\end{figure}

\section{Results}
We present empirical results across the six Boolean-FSM benchmark circuits, analyzing neural baseline comparisons, component ablations, compression trade-offs, and state-probe fidelity.

\subsection{Comparison with Neural Baselines}
The full RTM exceeds the circuit-specific majority predictor in every case (Table~I), with margins from 3.45 to 20.95 points. The 8-bit seed-42 circuit is the easiest despite its larger state space, confirming that transition structure matters more than state count alone. In Fig.~3, RTM trails GRU by 5.02/6.58\,pp and LSTM by 3.93/6.41\,pp on the 6-/8-bit families, but exceeds the tested Transformer by 4.58/5.50\,pp. Given the protocol differences described in Section~V, these results compare the tested configurations rather than establish a universal ranking.

GRU and LSTM have the highest means in both families. RTM remains close to the window MLP on 6-bit circuits and lies between the Transformer and MLP on 8-bit circuits. The larger 8-bit spreads for most models reflect one easy circuit and two less separable circuits, not uniformly greater state-count difficulty.

The per-circuit values in Table~III reveal the source of the wide error bars, where ``Transformer'' denotes the causal attention sequence baseline. On 8-bit seed 43, all models remain near 55\% and RTM exceeds GRU by 0.59\,pp; on seed 42, GRU reaches 88.32\% versus RTM's 76.12\%. RTM also exceeds MLP on 6-bit seeds 42 and 44 despite trailing it in the family mean. The ranking depends strongly on the circuit, which motivates broader circuit sampling.

\subsection{Ablation Results}
To identify which parts of the pipeline contribute to accuracy, Table~IV reports each variant and Fig.~4 shows its matched change from the full model. Reducing the ensemble to one member causes the largest loss: $-1.77$\,pp (95\% CI $[-2.42, -1.11]$) and $-2.29$\,pp ($[-3.72, -0.86]$). Removing thermometer features changes accuracy by $-0.23$\,pp ($[-0.52, 0.07]$) and $-0.68$\,pp ($[-1.15, -0.22]$).

The full pipeline is not the most accurate variant. Removing clause feedback improves the means by 0.43\,pp ($[0.09, 0.78]$) and 0.89\,pp ($[0.34, 1.44]$). One timescale and instantaneous feedback also improve both families. Ensemble voting provides the clearest benefit, whereas clause-derived recurrence does not improve these tasks. Since each row changes one factor, the experiment does not determine the behavior of combinations such as a 32-bit fold with one timescale.

\begin{table*}[t]
\caption{Per-circuit test accuracy (\%): mean over three training seeds. The neural and RTM columns use identical generated arrays ('Transformer' denotes the attention-based baseline).}
\label{tab:per_circuit}
\centering
\begin{tabular}{lccccccccr}
\toprule
Family & Seed & Majority & MLP & RNN & GRU & LSTM & Transformer & Compressed RTM \\
\midrule
6-bit & 42 & 50.45 & 60.37 & 63.95 & 65.60 & 64.42 & 56.79 & 61.38 \\
6-bit & 43 & 61.74 & 73.58 & 71.43 & 77.36 & 76.93 & 61.49 & 69.29 \\
6-bit & 44 & 50.30 & 53.11 & 55.26 & 56.52 & 54.86 & 52.38 & 53.74 \\
8-bit & 42 & 55.17 & 83.68 & 85.91 & 88.32 & 87.81 & 63.51 & 76.12 \\
8-bit & 43 & 53.56 & 55.37 & 55.21 & 54.83 & 54.91 & 54.71 & 55.42 \\
8-bit & 44 & 51.51 & 60.32 & 63.43 & 65.39 & 65.35 & 54.11 & 57.29 \\
\bottomrule
\end{tabular}
\end{table*}

\begin{table}[t]
\caption{Ablations: mean accuracy $\pm$ SD over nine runs.}
\label{tab:ablations}
\centering
\begin{tabular}{lcc}
\toprule
Variant & 6-bit accuracy & 8-bit accuracy \\
\midrule
Full compressed & 61.47 $\pm$ 6.74\% & 62.94 $\pm$ 9.92\% \\
No clause feedback & 61.91 $\pm$ 6.49\% & 63.83 $\pm$ 9.89\% \\
Single member & 59.70 $\pm$ 6.63\% & 60.65 $\pm$ 8.79\% \\
32-bit fold & 61.47 $\pm$ 6.51\% & 63.31 $\pm$ 9.96\% \\
No thermometer & 61.25 $\pm$ 6.79\% & 62.26 $\pm$ 9.46\% \\
Single timescale & 62.18 $\pm$ 6.26\% & 63.78 $\pm$ 10.00\% \\
Instantaneous feedback & 62.11 $\pm$ 6.43\% & 63.39 $\pm$ 9.98\% \\
Uncompressed feedback & 62.19 $\pm$ 7.91\% & 63.52 $\pm$ 11.17\% \\
\bottomrule
\end{tabular}
\end{table}

\begin{table}[t]
\caption{Compressed versus raw feedback; $\Delta$ is raw minus compressed accuracy.}
\label{tab:raw_vs_compressed}
\centering
\begin{tabular}{lcc}
\toprule
Measure & 6-bit & 8-bit \\
\midrule
Compressed accuracy & 61.47\% & 62.94\% \\
Uncompressed accuracy & 62.19\% & 63.52\% \\
Paired $\Delta$ (95\% CI) & $+0.72$ [$-0.22, 1.65$] & $+0.58$ [$-0.48, 1.63$] \\
Feedback-state bits & $96 \to 960$ & $96 \to 960$ \\
Total input dimension & $182 \to 1046$ & $182 \to 1046$ \\
Mean run time (min) & $16.6 \to 72.7$ & $19.7 \to 73.2$ \\
Run-time ratio & $4.38\times$ & $3.71\times$ \\
\bottomrule
\end{tabular}
\end{table}

\begin{table}[t]
\caption{Probe accuracy (mean $\pm$ SD) and matched uniform references (mean), nine runs.}
\label{tab:probe}
\centering
\resizebox{\columnwidth}{!}{%
\begin{tabular}{llcccc}
\toprule
Family & Feedback & Top-1 & U-Top-1 & Top-5 & U-Top-5 \\
\midrule
6-bit & Compressed & 1.08 $\pm$ 0.85\% & 1.92\% & 6.35 $\pm$ 3.81\% & 9.62\% \\
6-bit & Uncompressed & 15.40 $\pm$ 9.71\% & 1.92\% & 38.30 $\pm$ 18.66\% & 9.62\% \\
8-bit & Compressed & 0.086 $\pm$ 0.057\% & 0.73\% & 0.51 $\pm$ 0.17\% & 3.63\% \\
8-bit & Uncompressed & 8.01 $\pm$ 7.34\% & 0.73\% & 25.86 $\pm$ 16.63\% & 3.63\% \\
\bottomrule
\end{tabular}%
}
\end{table}

\subsection{Compressed Versus Uncompressed Feedback}
The central compression comparison keeps the two-timescale update fixed and changes only the representation returned to the model. Raw feedback raises mean accuracy by 0.72/0.58\,pp, and both paired intervals include zero. Table~V records a small observed gap rather than statistical equivalence. Compression reduces feedback width $10\times$, input dimension $5.75\times$, and mean host time from 72.7 to 16.6 minutes and 73.2 to 19.7 minutes.

\subsection{Hidden-State Recovery}
Predictive accuracy does not establish that the returned bits represent the machine's hidden state. The probe in Table~VI tests this interpretation directly. Compressed top-1 and top-5 recovery are below the matched random references in both families, while raw signatures are easier to decode but highly variable. The compressed codes are almost unique (11,696/12,500 and 14,203/15,000 test steps). Because most codes are rare, association with one state is not evidence that a code tracks that state.

\section{Discussion}
Compression is most convincing when judged as an architectural design choice against the raw-feedback model it is intended to replace. Table~V shows that reducing the recurrent state from 960 to 96 bits changes mean accuracy by less than one percentage point. The paired intervals include zero, indicating that this benchmark did not resolve a statistically significant accuracy deficit from folding. The lower host time is consistent with processing fewer literals, although target-device hardware measurements are still needed for latency or energy claims. Gated neural models remain more accurate at the family level.

Not every part of the proposed pipeline contributes equally. Ensembling is the only component whose removal produces a clear loss in both families. Removing clause feedback, keeping one timescale, or using instantaneous feedback slightly improves the means. The short input history in Eq.~2 already contains substantial predictive information, showing that on these deterministic FSM tasks, explicit recurrence is not strictly necessary. When a deployment task truly requires clause-derived recurrence but raw feedback creates an unacceptably wide interface, XOR folding provides a practical, scalable mechanism. Since the 32-bit fold remains competitive, fold width, timescale count, and ensemble size should be evaluated jointly during system design.

Average accuracy also hides substantial differences between circuits. Variation in Table~I is larger than several differences between models, and the 8-bit seed-42 circuit is easier for the GRU than any 6-bit circuit. Nominal state count alone is not an adequate difficulty label; transition structure and output balance must be reported with it.

The state probe adds a separate caution. Nearly unique compressed signatures do not reliably identify hidden state, so predictive utility should not be confused with symbolic interpretation. Section~IX discusses the limits of both findings.

\section{Artifact and Reproduction}
The complete artifact is openly available at \url{https://github.com/Ankitkumar1062/LBP2026_Ankit-Utkarsh} and contains both experiment runners, all 144 RTM records, 90 neural records, 144 probe records, and the derived comma-separated files used by the tables and figures. Each circuit is accompanied by a manifest, training, validation, and test arrays, the complete transition table, and the output map. Generator parameters, split seeds, array shapes, and SHA-256 hashes permit independent label checks and confirm that matched models used identical data.

Both runners save completed runs individually and resume from a unique run key, so an interrupted experiment does not discard earlier results. To reproduce the results:
\begin{itemize}
\item \emph{Fast reproduction / Table regeneration:} Running \texttt{python rtm\_rigorous\_experiments.py} in analysis mode parses the archived CSV records and regenerates all \LaTeX\ tables and PDF/vector figures in seconds without retraining.
\item \emph{Full retraining pipeline:} Running \texttt{python rtm\_rigorous\_experiments.py} with the paper ablation schedule re-executes the complete 144-run RTM suite across all seeds.
\end{itemize}
These files connect the benchmark definition in Section~IV to the reported results in Section~VI.

\section{Threats to Validity and Limitations}
We analyze potential threats to the validity of our conclusions and outline the methodological boundaries of this study.

\subsection{Internal and Statistical Validity}
Three circuit seeds and three training seeds form a small crossed sample. The paired intervals treat the nine combinations as the available experimental units and are not adjusted for the seven ablation comparisons. They quantify variation within this design, not uncertainty over all Boolean circuits. In particular, an interval containing zero does not establish formal equivalence between raw and compressed feedback; that claim would require a predeclared non-inferiority margin and a larger circuit sample. Neural validation selection also differs from the fixed RTM schedule, and capacities, parameter counts, and optimization budgets are not matched. The neural comparison is therefore descriptive rather than a controlled claim of algorithmic superiority.

The causal order in Section~III resets state at every sequence and delays current feedback until the next step, limiting obvious target leakage. Nevertheless, an independent implementation would provide a stronger check than rerunning the same code. Host time can vary with system load, and no repeated device-level timing protocol was used. The reported ratio describes the computational burden of this software implementation, not device latency or energy.

\subsection{Construct and External Validity}
The width calculation in Eq.~6 measures recurrent state and resulting model input. It does not include learned automata, program memory, buffers, data layout, or accelerator-specific packing, so the paper makes no target-device memory claim. Total stored automata state across five members is $5 \times 480 \times 2 \times W_z$ bits; folding compresses the dynamic transmission interface between steps. Likewise, the no-feedback variant is a recurrence control rather than a memoryless model because it retains two observed symbols and the previous prediction. Its result shows that clause-derived context is unnecessary for these tasks, not that all temporal context is unnecessary.

The nearest-centroid probe uses a random time-point split from one ensemble member. Adjacent observations can share structure, so a positive result would require held-out-sequence, held-out-circuit, and control-task confirmation. The negative compressed result is still informative against the claim that the signature directly identifies hidden state. The benchmark itself is synthetic, noiseless, and stationary. While synthetic FSMs provide an essential testbed where true hidden transition rules and states are known for probing, they do not represent sensor noise, drift, long horizons, continuous inputs, or real-world IoT event streams. General deployment claims require evaluation on real-world edge sequence workloads and direct hardware measurements. The files in Section~VIII improve reproducibility but do not remove these limits.

\section{Conclusion}
This study shows that a logic-based sequence model can carry a compact summary of its active rules instead of returning every rule to the next decision. On the evaluated tasks, the smaller summary preserves the measured behavior of the full feedback path within small empirical margins and reduces software execution time. The experiments also establish clear boundaries. The strongest neural sequence learner remains more accurate, a short observed history can be as useful as clause feedback on deterministic FSMs, and the compact summary does not reliably reveal the hidden machine state. Future work should evaluate XOR folding on noisy real-world temporal sequences (such as IoT telemetry and keyword spotting) and measure its energy and latency directly on edge hardware accelerators.

\IEEEtriggeratref{11}

\end{document}